# An Analysis and Comparison of ACT-R and Soar

**John E. Laird**    LAIRD@UMICH.EDU
Computing Science and Engineering, University of Michigan, Ann Arbor, MI 48109-2121 USA

## Abstract

This is a detailed analysis and comparison of the ACT-R and Soar cognitive architectures, including their overall structure, their representations of agent data and metadata, and their associated processing. It focuses on working memory, procedural memory, and long-term declarative memory. I emphasize the commonalities, which are many, but also highlight the differences. I identify the processes and distinct classes of information used by these architectures, including agent data, metadata, and meta-process data, and explore the roles that metadata play in decision making, memory retrievals, and learning.

## 1. Introduction

Cognitive architectures, such as ACT-R (Anderson, 2007) and Soar (Laird, 2012), are designed to provide the computational infrastructure sufficient for supporting the knowledge representations and processing of cognitive models and AI agents. In this paper I develop a simple and sufficient taxonomy for the types of information that are represented and processed in these architectures. Using this taxonomy, I analyze and compare the current versions of ACT-R (Bothell, 2020) and Soar (Laird et al., 2017) at a level of detail beyond prior analyses. I choose ACT-R and Soar because of my familiarity with them, their prominence, and their similarity. Although similar in structure, they have been shaped by different primary goals: cognitive modeling of human behavior for ACT-R and development of general AI agents with complex cognitive capabilities for Soar. Conveniently, their secondary goals overlap with their differing primary goals: ACT-R has been used for real-world agents and Soar has been used for modeling human behavior.

This analysis extends the original analysis that led to the "Common Model of Cognition" (Laird, Lebiere, & Rosenbloom, 2017). The Common Model is an abstract theory of the mind that draws from research across cognitive science, AI, and neuroscience, inspired by Newell's work on Unified Theories of Cognition (Newell, 1990). A unique contribution of this analysis is examining the interaction between data and metadata, including how metadata are maintained by the architecture and the role they play in architectural processing for memory retrieval, decision making, and learning. Other analyses and comparisons of cognitive architectures and similar frameworks have focused on defining the space of cognitive architectures (Langley, 1983), analyzing the general types of knowledge and processing (Jones & Wray, 2006), or providing a broad overview of current and past cognitive architectures (Kotseruba & Tsotsos, 2020). In contrast, this is a detailed comparison of the specific representations and processes used in two cognitive architectures, with the goal of improving our understanding of these Common Model architectures. An equally important goal is to identify regularities in the design and interactions of their constituent modules and underlying data and metadata representations that can inform the analysis and design of other cognitive architectures.

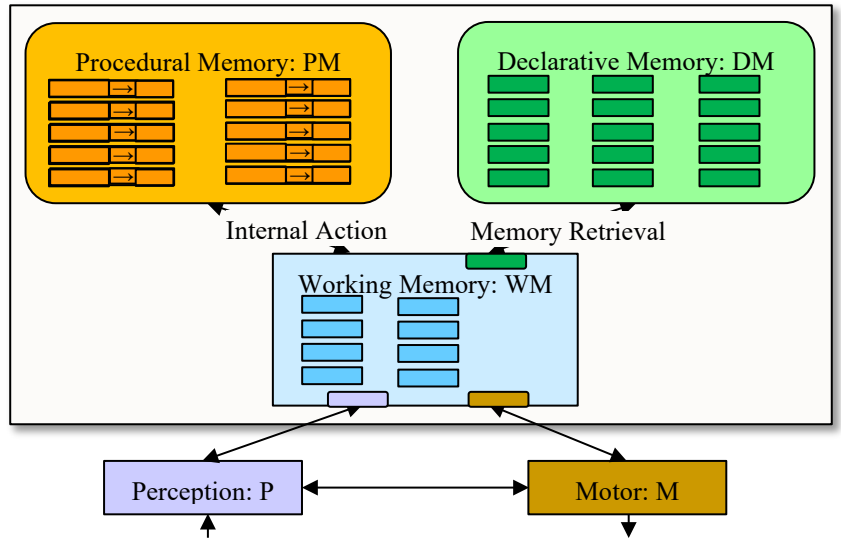

Figure 1: Modules of the Common Model of Cognition.

Section 2 introduces the organization, agent data, metadata, and processing of these cognitive architectures, defining terminology and concepts, which are then used in Sections 3-5 to compare and contrast the details of working memory, procedural memory, and declarative memory in ACT-R and Soar.[1] Each section describes commonalities shared by the two architectures, followed by those aspects that differentiate the architectures from each other.[2] For the casual reader, I recommend reading Section 2, then skimming over Sections 3-5, reading those sections and subsections of specific modules of interest. Section 6 summarizes the analysis, highlighting the commonalities and differences of these architectures, and discussing other types of data and metadata that should possibly be considered for inclusion in future cognitive architectures.

## 2. Overview of Architecture: Data and Processes

A cognitive architecture defines the task-independent memory systems, data representations, processes, and interfaces of an intelligent agent. Figure 1 shows the memories, processes, and their interconnections as characterized by the Common Model,[3] which are shared by ACT-R and Soar. That structure, in terms of the modules and their connections, is fixed and cannot be modified by learning. Starting from the lower left, perception (P) delivers changes to working memory (WM). Working memory contains the agent's current representation of the environmental situation, augmented with its interpretation of the situation, as well as the agent's goals, intentions, and other data that drives the agent's immediate behavior. It includes *buffers* for interacting with the other modules. Procedural memory (PM) contains the agent's knowledge about how to act, such as skills and procedures. Accessing procedural memory entails selecting and executing of a single *internal action* based on the contents of working memory. Many cognitive architectures use rule-like structures to represent procedural knowledge because they independently encode when an action can be performed (the conditions which test working memory) and how to perform it (the actions that modify working memory). The changes made by an internal action drive the overall cognitive cycle by initiating processes in the modules for:

1. the selection and execution of a subsequent internal action from PM;
2. a retrieval from declarative memory (DM);
3. execution of a motor action (M);
4. modulation of perception (P).

---

[1] Perception and motor are not included because there is little to distinguish them between the two architectures.

[2] A formalization is available at http://laird.engin.umich.edu/wp-content/uploads/sites/331/2021/08/formal-3.pdf.

[3] The link from motor to WM was not in the original Common Model, but both ACT-R and Soar include the ability of motor to provide feedback to WM during the execution of a motor action.

Declarative memory (DM) contains long-term facts, beliefs, and experiences, which can be retrieved into working memory by the creation of a cue in the DM buffer. The individual modules (P, PM, DM, M) run independently and asynchronously, with their own memory structures and processing (including learning). Complex cognition arises from sequences of these primitive cycles, with no additional modules for planning, language processing, or metacognition.

## 2.1 Agent Data

Data in the memories consist of *memory elements* or just *elements*. An element is a symbol structure that can be independently tested, created, or deleted. It can contain symbols, numbers, and strings. *Agent data* are the elements in WM, DM, and PM. They encode the agent's knowledge about itself, its skills, it tasks, and the world. There are three classes of agent data that are differentiated by the processes that create, test, and modify them.

1. **Internal agent data.** These are short-term symbol structures that are created by P, PM, or retrievals from DM in WM, or long-term structures created by learning processes in PM or DM.
2. **Module commands.** These short-term elements are created by PM actions and are sent to a module to initiate its processes, such as a retrieval from DM or a command to M. There is an innate/fixed set of symbols that define the commands for each module.
3. **Module status data.** These elements are created by a module in its WM buffer, providing feedback on the module's processes, such as that a retrieval from DM was successful. These are the one source of meta-process data – data about the underlying architectural processing.

With few exceptions, architectural processing of agent data is sensitive only to the *form* of agent data (that agent data elements are graph structures or rules), but not its content (the specific symbols, numbers, or strings). Exceptions are module commands and module status data, which include innate symbols with fixed semantics. The meaning of agent data is determined by how the agent derives them from perception (grounding), how they connect with other agent data, and how the agent reasons with them using other data and uses them to interact with the world – a procedural semantics view of meaning.

## 2.2 Agent Metadata

In addition to agent data, a cognitive architecture maintains agent metadata,[4] which are data *about* agent data. These influence how agent data are processed by the architecture. Agent metadata are created and updated exclusively by the architecture in parallel with, and as a side effect of, the processing of agent data. Metadata are often statistical/numeric but can be relational/symbolic. They can be associated with individual data elements or collections of memory elements. *Base-level activation* (BLA), or just *activation*, is a common type of metadata, originally defined in ACT. It encodes the recency and frequency of creation and usage of an element and is used to represent the likelihood that an element will be useful in the future.

In ACT-R and Soar, metadata are used exclusively by architectural processes for memory retrievals, decision making, and learning. They are not represented in working memory and they cannot be tested or directly modified by agent data processing. Thus, there is a "wall" between agent data and agent metadata. An agent can access metadata indirectly by monitoring its own behavior, such as testing the speed of DM retrieval (Anderson et al., 2021), and influence them indirectly through deliberate actions such as rehearsal. In AI, metadata sometimes refers to any data used in metalevel reasoning/processing, such as during deliberate self-reflection, even if it is agent data, which is a broader definition than used here.

## 2.3 Processing

Every module (P, PM, DM, M) has specific associated processes. Perception has processes that transduce input data from sensors. PM has processes for selecting and applying a single internal action, whereas DM has processes for retrieving data from its memories. PM and DM also have processes for learning new elements. Motor has processes that convert commands to action in the environment. Metadata processing is a side effect of these processes over agent data and all changes to agent data and metadata occur exclusively through these processes. DM retrievals (limited to one at a time) and M actions are initiated by the creation of cues or actions requests by PM in an associated buffer. Sharing of data and metadata is limited to communication through WM (except for a direct connection between P and M).

## 3. Working Memory

In cognitive science, working memory can refer to elements in both short-term and long-term memories that are potentially relevant to the current situation. For our purposes, WM is restricted to those memory elements that can cue selection of actions from procedural memory. These elements can be the result of perception, internal PM actions, retrievals from long-term DM, and module status data, such as feedback from M. They can include information about the current situation, the agent's current goals and plans, and any other elements that drive behavior.

---

[4] Metadata has sometimes been called "subsymbolic" information. I avoid that term as it can refer to information used to *implement* symbolic representations and processes, as in neural-based symbol processing.

## 3.1 Working Memory Data

In both ACT-R and Soar, working memory contains relational graph structures. An individual element consists of a triple: a node, a labeled edge, and a value. The node and edge labels are symbols. The value can be another node, a primitive symbol, string, or a number. Elements that share a common node (not a value) are collectively called a *chunk*,[5] where the node symbol is also called the *chunk name*. The distinction between individual elements and chunks (collections of elements) is important in how data are processed.

Working memory includes designated areas called *buffers* that provide interfaces to other modules, including perception, declarative memory, and motor. To support interaction with the associated module, the following six fields can be part of a buffer:

- A chunk generated by perception;
- A command to be sent to a module, such as for performing a motor action;
- A chunk to be sent as part of a motor action;
- A cue for a retrieval from the DM module, which is a partial specification of a chunk;
- A chunk retrieved from declarative memory;
- The status of the processing in a module, such as "busy," "success," or "failure."

Procedural memory actions can test any working memory data and modify any working memory element, but they cannot modify module status data, which is created by the architecture.

### 3.1.1 ACT-R Working Memory Data

In ACT-R, working memory corresponds to a *fixed* set of modules with associated buffers, each of which contains a field for holding a chunk (such as the retrieval field for the DM buffer). These buffers are the only memory structures that can be tested and modified by procedural memory. In ACT-R terminology, a chunk consists of a unique identifying *name* and an unordered set of labeled *slots* with *values*. The value of a slot can be a chunk name, a string, a number, or a distribution over numbers. The chunk name, together with a single slot-value pair, corresponds to a single memory element as defined above. Each element can be independently created, tested, and deleted. Storage and retrieval to DM are performed on chunks and not individual elements.

The DM module has a single field for requesting a retrieval and then receiving a retrieved chunk, so that the retrieved chunk overwrites the request. The buffers associated with DM, P, and M contain module-specific status data that can be tested but not created or modified by PM. Module commands are encoded directly as actions of PM elements (called productions in ACT-R) and sent directly to the module without being represented in the buffer. This contrasts with Soar, where module commands are placed in the command field of the buffer.

---

[5] In this paper, I use ACT-R terminology to describe WM structures, so that even though a chunk in Soar is a learned rule, in this paper, *chunk* refers to a WM structure.

For a given agent, there is a fixed set of modules and buffers, but the number varies across agents depending on the task that an agent is designed for. Agents that involve external interaction include DM, P, M, a goal buffer, and an imaginal buffer. Examples of other task-independent modules (with associated buffers) that have been developed in ACT-R research include a temporal module for estimating time and a controller module for learning settings of control parameters (Anderson et al., 2021). Many ACT-R agents also include task-specific modules to provide additional buffers to increase intermediate storage or to augment the perceptual and motor systems. For example, an ACT-R agent that processes natural language may have specialized buffers for intermediate results in language processing (Ball, 2011).

### 3.1.2    Soar Working Memory Data

In Soar, working memory is a connected graph structure rooted in a node called the *state*. An individual element consists of a symbol called an *identifier* (which is a node – equivalent to a chunk name), an *attribute* (the edge/slot), and a *value*. Values can be identifiers, numbers, or strings. Thus, an element is a labeled edge between two nodes (the identifier and the value) in a graph, corresponding to a slot-value pair of a chunk in ACT-R. A collection of elements that share an identifier is equivalent to an ACT-R chunk. For consistency, I use *element* to refer to a single slot-value triple of a chunk, and *chunk* to refer to elements that share the same node, even when referring to Soar memory structures. Many operations apply to individual elements, but some, such as a retrieval from declarative memory, apply to a chunk as a whole.

In Soar, WM elements are not restricted to buffers and there is no limit to the number of WM elements. However, all elements must be *linked*, possibly indirectly, to a state, so that there is a path through the graph from a state to each WM element. Buffers are a part of the state and are used for interactions with DM, P, and M, with fields for commands, cues, retrievals, etc.

The WM of a Soar agent is initialized with a single *top state* and all interactions with the environment occurs through P and M buffers in that state. When there is an impasse in procedural memory processing, a new substate is automatically created in working memory. A substate is a chunk consisting of elements that describe the reason for the impasse, including an element that links to the superstate (which is maintained in WM in addition to the substate).Thus, a substate provides module status data created when PM fails to select and apply an internal action, providing an entry for metacognitive reasoning. When the impasse is resolved, the substate is automatically removed.

### 3.1.3    Analysis of ACT-R and Soar Working Memory Differences

ACT-R's and Soar's working memory have the same functional role and the buffer structure for supporting module interaction. Although the details of the buffers vary, both have read-only status fields for receiving feedback from modules. The most important difference is that in ACT-R, WM is restricted to a fixed set of buffers, whereas in Soar, there is no constraint on the breadth or depth of the graph structure. The limits on WM in ACT-R leads to extensive use of declarative memory to hold intermediate structures.

For real-world application, ACT-R's limits lead to the inclusion of additional task-specific buffers. Soar's WM can make it easier to develop complex AI agents, but also makes it less likely to model aspects of human reasoning related to short-term memory management. Soar's substates support metareasoning about the current situation, which enable many types of complex cognition (planning, perspective taking, hierarchical task decomposition). Metacognition has been relatively unexplored in ACT-R.

## 3.2 Working Memory Metadata

There are multiple types of metadata that are associated with a WM element, all automatically maintained by the architecture in parallel with agent data processing. Two generic types of WM metadata are used by both ACT-R and Soar:

- **Copy-of**: When a retrieval from DM is made, a copy of the selected DM chunk is made and installed in the DM buffer, which I refer to as *WDM*. A WDM has relational metadata that indicates its source chunk in DM so that the architecture can maintain a correspondence between the two for when the WDM is tested or modified.
- **Derivation**: To support procedural learning, metadata are maintained about agent data used to *derive* an element. If an element is created by procedural knowledge, a copy of the instantiated procedural element is saved.[6] If an element is created via DM retrieval, the WM elements that served as the cue are saved. Metadata are not maintained for module status and perception-based elements because they are not incorporated in learning.

### 3.2.1 ACT-R Working Memory Metadata

ACT-R maintains copy-of metadata for WMD chunks as described above. Instead of explicitly maintaining derivation metadata, it maintains a memory of the last fired production instantiation, which is used in procedural learning (Section 4.3). That memory essentially encodes derivational metadata for all actions of that instantiation so that during learning, they can be merged with conditions of the production that fires next.

### 3.2.2 Soar Working Memory Metadata

Soar also maintains the metadata described above, but also stores two additional types of WM metadata:

- **Activation**: This is the quantitative base-level activation described earlier, where metadata are associated with individual WM elements (except module status and perceptual data). Activation is computed from when a WM element was created and accessed. It biases retrievals from DM and can be used in automatic WM forgetting.

- **Substate Level:** Each WM element includes relational metadata that encodes the highest substate to which an element is linked. These are used to determine the results of subgoals and are also used in PM learning.

---

[6] A PM instantiation includes the bindings of variables in its conditions and actions to WM elements (see Section 4).

### 3.2.3    Analysis of ACT-R and Soar Working Memory Metadata Differences

ACT-R and Soar have similar qualitative metadata in terms of functionality for WM elements. Activation of Soar's WM elements is rarely used for WM forgetting, but it is used for biasing retrievals from DM. Substate level metadata in Soar is necessary to support substates in which it can pursue metareasoning.

### 3.3 Working Memory Forgetting

ACT-R does not have a forgetting mechanism for working memory because it has a fixed working memory consisting of the buffers whose contents are overwritten as new information becomes available. Soar automatically removes substate WM elements when the substate terminates because those WM elements are no longer accessible. Soar has the capability to remove WM elements when their activation falls below a fixed threshold, but this is rarely used as agent reasoning usually removes elements that are no longer relevant.

## 4.    Procedural Memory

Procedural memory contains knowledge about skills and reasoning: when and how to perform internal and external actions. We first discuss the representation of PM elements and PM retrieval (the main process associated with PM), and then discuss PM metadata.

### 4.1 Procedural Memory Data and Usage

A PM element has a rule-like structure with two parts, conditions and actions, distinguishing it from WM and DM elements. The conditions tests WM elements to determine when it is possible to execute the actions, while the actions create and remove individual WM elements. The selection is based on a combination of factors, including the comparative utility of competing PM elements that match WM. The selection and application of a single PM element drives the cognitive cycle by making one or more changes to working memory. Those changes can lead to further execution of PM actions, DM retrievals, initiation of motor actions, or interactions with P. As a side effect of a PM's actions, the metadata of relevant WM, PM, and DM elements are updated. There is no consensus as to the exact structure of a PM element – whether it is a rule (as in ACT-R) or a collection of rules (as in Soar). However, there is agreement that it is the primitive unit of conditional deliberate action, which in modeling human behavior, takes ~50 msec to select and apply.

### 4.1.1    ACT-R Procedural Memory Data and Retrieval

ACT-R's PM elements are individual production rules whose conditions are matched against chunks in buffers. A single rule is selected based on degree of match to WM elements, combined with the rule's utility, a type of metadata described below. A rule action can create new chunks in buffers, modify individual elements in buffer chunks, or remove (clear) existing chunks from buffers. ACT-R does not report the status of PM retrievals in any buffer.

### 4.1.2    Soar Procedural Memory Data and Retrieval

For this analysis, a PM element in Soar is an *operator*. An operator consists of a collection of rules that propose, evaluate, and apply the operator. Rules fire in parallel, so multiple operators can be proposed and evaluated simultaneously. A single operator is selected based on the results of proposal and evaluation. Once an operator is selected, application rules for that operator fire in parallel, creating or deleting individual WM elements. Although Soar operator elements are represented in WM, they should not be considered agent data because they are memory structures used exclusively for selecting and applying an operator. Instead, they should be considered data structures that are internal to accessing procedural memory.

Soar's PM also includes *elaboration* rules, which fire in parallel and create elements in WM when their conditions match and then retract those elements when their conditions no longer match. These rules (and the data they produce) are not associated with specific operators but expand the set of available data used during operator selection and application (PM retrieval), acting as a preliminary stage in PM retrieval.

This finer-grained representation of procedural knowledge means that the process of retrieving and applying that knowledge can fail for multiple reasons: no operator is proposed, multiple operators are proposed but there is insufficient knowledge to choose among them, and an operator is selected, but there is insufficient knowledge to apply it. All of these failures lead to an *impasse* in Soar and the creation of a new WM chunk, called a *substate,* which includes the type of impasse. This is how PM module status information is represented in Soar. Once the substate is created, the process of selecting and applying operators recurs until the impasse is resolved, which leads to the removal of the substate and associated WM elements.

### 4.1.3    Analysis of ACT-R and Soar Procedural Memory Data

Soar provides a finer-grained representation of procedural knowledge, using multiple rules to represent a PM element, as opposed to a rule in ACT-R. This leads to multiple tradeoffs. With the more coarse-grained representation, ACT-R models retrieve only one rule and learn only one rule. With a more fine-grained representation, Soar models can have independent knowledge for proposing, selecting, and applying a PM element, which they can learn independently. Also, Soar has a richer language for expressing the knowledge used to select among competing PM elements. In ACT-R, there is utility (described below), which is also in Soar, but there are also preferences that allow preferring one element over another, making an element a default to be selected only if there are no other elements, rejecting elements, and making an element preferred to all other except those that have a similar preference. Soar's finer-grain representation, together with the detection of impasses, allows for the automatic creation of subgoals within which hierarchical problem solving and metareasoning can occur.

## 4.2  Procedural Memory Metadata

Procedural *utility* is associated with individual PM elements and biases selection from PM. It is computed from *reward*, which is neither agent data nor agent metadata. Utility is updated by a form of temporal difference learning (Sutton, 1988). In both architectures, reward is computed by rules or by internal architectural computations (such as for intrinsic reward or reward derived from the environment). Computing reward via rules is convenient and flexible, especially for internal goals, but it is problematic when there are novel tasks that require new reward rules.

### 4.2.1 ACT-R Procedural Metadata

In ACT-R, when a reward is created, utility metadata of all rules that have fired since the last reward are updated. To support this, PM elements maintain addition metadata, including the time of the last reward and the time of previous PM element firing.

### 4.2.2 Soar Procedural Metadata

In Soar, some evaluation rules (called RL rules) have associated utility metadata that are updated using temporal difference learning. Soar also includes activation metadata for learned rules that is used for forgetting learned rules that are rarely used (Derbinsky & Laird, 2013).

## 4.3 Procedural Memory Learning

The updating of the utility metadata described above supports reinforcement learning in both architectures. A second type of procedure learning involves the creation of new rules that summarize multiple PM selections and executions by composing them using derivational metadata. This type of learning compiles away intermediate processing, including retrievals from DM, but not environmental interactions, speeding future reasoning. The details of this type of learning differ between the architectures.

### 4.3.1 ACT-R Procedure Memory Learning

In addition to reinforcement learning, ACT-R learns new rules by *production compilation*, which is performed between pairs of rules that fire in sequence, eliminating the actions of the first rule that are tested in the second rule. This eliminates any DM retrieval in the first rule whose result was tested in the second. Compilation is limited so that the combined computation is deterministic. The new rule is initialized with low utility, so it will not immediately replace the original pair. However, when the new rule duplicates an existing rule, its utility is boosted. With enough repetitions, the new rule will be selected in place of the original rules.

### 4.3.2 Soar Procedure Memory Learning

In addition to reinforcement learning, Soar also learns new rules by combining existing rules, using a process called *chunking*. Chunking composes rule instantiations that fire in subgoals and ultimately produce a result. The process is initiated when a result is created in a substate, followed by composing the rule instantiations and DM retrievals that occurred in the substate and that led to the result. Instead of replacing a pair of existing rules, the new rule bypasses the creation of an impasse and the reasoning that would occur in the ensuing substate.

## 4.4 Procedural Memory Analysis and Discussion

Although ACT-R and Soar differ in the underlying representation of procedural memory elements (rules vs. operators), the overall processing and core metadata are similar, including the support of two learning mechanisms, one that learns the utility of PM elements, and a second that compiles PM processing (and DM retrievals) to generate new, more efficient PM elements. ACT-R compiles over pairs of rules that fire in succession, whereas Soar compiles the processing in substates, which can involve arbitrary numbers of rules. Another difference is that in ACT-R, a rule must be learned multiple times for it to eventually be used, whereas in Soar, once a rule is learned, it does not need to be relearned. However, when we have attempted to closely model human procedure learning in Soar, we have found it necessary to modify the architecture so that learned rules must be learned multiple times before they are used (Stearns, 2021).

## 5. Declarative Memory

Functionally, declarative memory is the equivalent of a large database that contains facts and other long-term data. An important difference between DM and WM is that DM chunks must be retrieved into WM before they influence reasoning through PM actions. WM contains a DM buffer that includes fields for cuing a retrieval, receiving the results of a retrieval, and module status. The two DM processes are retrieving items into WM (5.3) and learning new DM structures (5.4). There are no explicit forgetting mechanisms for DM, although DM chunks with sufficiently low base-level activation will never be retrieved.

## 5.1 Declarative Memory Data

DM uses the same representations as WM with elements organized into chunks.

### 5.1.1 ACT-R Declarative Memory Data

ACT-R has a single declarative memory as described above and has two buffers associated with DM in WM. One is for retrieving a single chunk from DM. The other buffer is for retrieving matched chunks are "blended." The status field includes whether a retrieval was requested, whether it succeed or failed, and whether the memory system is in the process of performing a retrieval.

### 5.1.2 Soar Declarative Memory Data

Soar has two long-term declarative memories: semantic memory (SM) and episodic memory (EM), each with its own buffer in WM. They differ in the type of information they contain, how that information is added, and how it is retrieved. Semantic memory is similar to ACT-R's declarative memory in that it stores memory elements without temporal/sequential or relational metadata. Episodic memory contains episodes, which are "snapshots" of the complete top state of WM, but not substates created because of impasses.

## 5.2 Declarative Memory Metadata

Metadata are associated with both individual DM elements and DM chunks. Base-level activation is associated with chunks, which then influences retrievals from DM (Section 5.3). ACT-R and Soar differ on the exact events that contribute to this activation. Activation metadata can also be associated with individual elements or pairs of chunks and is often called *association strength*. These metadata bias the spread of activation to connected chunks during a retrieval. Once again, ACT-R and Soar differ on which events contribute to activation, but in both cases element activation metadata influences activation spread.

### 5.2.1   ACT-R Declarative Memory Metadata

ACT-R maintains activation metadata for chunks in DM, which are updated whenever a chunk is store/restored into DM. Recently, associational metadata between chunks has been added back to ACT-R, which provides information about the recency and frequency with which chunks co-occur in WM (Anderson et al., 2021).

### 5.2.2   Soar Declarative Memory Metadata

In Soar's semantic memory, metadata are associated with both chunks and elements. Activation metadata of chunks are updated when a chunk is stored or when it is retrieved. Activation metadata for elements are updated when a corresponding WDM element is tested or created. Episodic memory elements have temporal metadata unrelated to activation. The metadata includes the time at which an element was added to and deleted from episodic memory. Relational metadata are associated with each episode that allows an agent to retrieve temporally previous or next episodes.

## 5.3 Declarative Memory Retrieval

To retrieve memory elements from declarative memory, a cue is created in the declarative memory buffer in WM.[7] Only one retrieval is possible at a time. The cue provides a partial specification for a memory element in DM, which is searched to find the best matching chunk. Three types of metadata, as well as noise, influence retrievals: the degree of match of the cue to a DM chunk; the base level activation of DM chunks that are candidates for retrieval; and the activation that is spread from chunks in WM to DM chunks. Retrieval in ACT-R and Soar differ in the details, but in both cases retrieval is based on a cue that constrains which DM chunks can be retrieved, activation of chunks in DM, which biases the retrieval based on history (activation of individual DM chunks), and context (the result of spread for WDMs). In both, chunks recently retrieved from DM can be inhibited to avoid repeated retrievals of the same DM chunk. As with PM, metadata for DM and WM elements are updated as a side effect of a DM retrieval.

---

[7] ACT-R also supports spontaneous retrieval, as described in Section 5.3.1.

### 5.3.1    ACT-R Declarative Memory Retrieval

In ACT-R, the cue can include constraints on the values of slots, such as that a retrieved value must not equal a specific value or that a retrieved value must be above/below a numeric threshold. ACT-R supports two types of retrievals by using different buffers. One retrieves an existing DM chunk. The second "blends" the values of all chunks that match the cue using a weighted average of their activation (Vinokurov et al. 2013). In ACT-R, spreading activation initiates from the *values* of elements in buffers, and then spreads one level to chunks that share those values. ACT-R also supports spontaneous retrieval. If the retrieval buffer is empty and the most highly activated DM chunk exceeds a fixed threshold, that chunk is retrieved.

### 5.3.2    Soar Declarative Memory Retrieval

Soar biases retrievals from DM by including WM activation metadata of elements in the cue buffer, so that it prefers DM chunks that contain elements in the cue high activations. Activation spreads from DM chunks that are also in WM to other DM chunks, from a chunk through individual elements to chunks that are values. The spread can continue through multiple levels of additional DM chunks, controlled by a parameter. Soar provides the option of retrieving children of a result, recursively to a specified depth. Soar also supports *non-cued* retrieval from semantic memory using a chunk's name, which directly retrieves that chunk.

In episodic memory, the cue is a partial specification of WM and the most recent episode that fully matches the cue is retrieved. If no episode matches fully, the episode with the best match is retrieved, with ties broken by recency. Next or previous episodes of a retrieved episode can also be retrieved, which allows for recreation of an extended experience. Spontaneous retrieval from episodic memory has been used in experimental versions of Soar.

## 5.4 Declarative Memory Learning

As in PM learning, in DM learning there are two classes of learning mechanisms. The first are described in Section 5.2 and update DM activation metadata, influencing future retrievals. The second are mechanisms that copy or merge WM chunks into declarative memory. The exact details for performing those processes vary by architecture and memory module.

### 5.4.1    ACT-R Declarative Memory Learning

ACT-R learns activation metadata and associational metadata as described in Sections 5.2.1 and 5.3.1. ACT-R learns new DM knowledge by adding a chunk in a buffer to DM whenever a buffer is "cleared."

### 5.4.2    Soar Declarative Memory Learning

As described in Sections 5.2.2 and 5.3.2, Soar also learns activation metadata that influences retrievals. Soar does not have an automatic storage mechanism for semantic memory, and relies on explicit storage of chunks in WM into DM. For episodic memory, all changes to WM are stored automatically into EM.

### 5.5 Declarative Memory Analysis and Discussion

ACT-R declarative memory and Soar's semantic memory are similar long-term declarative memories, supporting similar mechanisms for retrieval using similar metadata, although the details differ, especially in how activation spreading occurs. ACT-R's declarative memory has been used for some aspects of episodic memory, but it does not directly support the automatic simultaneous storage of the contents of all buffers along with temporal relations to in a way that they can be retrieved in the future. Temporal information can be deliberately encoded in chunks, which incurs processing overhead during reasoning and does not capture the concurrent contents of multiple buffers.

## 6. Summary and Discussion

Well, that was a bit of a slog, but my hope is that this analysis leads to insights about these cognitive architectures and possibly others. Even though ACT-R is designed for cognitive modeling and Soar is designed for developing general AI agents, they have many commonalities:

- There is a basic cognitive cycle and a fixed set of core components (P, WM, DM, PM, M).
- There are a small number of basic types of information: agent data, agent metadata, and module status (meta-process data), and only a limited set of types of data having predefined meaning, including module status data and motor commands.
- Buffers are used for interfacing between WM and other modules.
- Module status data in buffers are the only source of meta-process data.
- Declarative knowledge for WM and DM is represented as symbolic graph structures.
- PM processing includes selecting and executing a single rule-like element using utility metadata. PM has two learning mechanisms – one for learning utility of PM elements and the other that learns new elements via knowledge compilation.
- DM processing involves retrievals and learning of chunks biased by metadata.
- Metadata are updated as a side effect of PM and DM processes. Metadata are not examinable or modifiable by agent data.

Although they are very similar, automatic (or even straightforward) conversions of programs written for one architecture to another are not possible. There are abstract languages that compile into both architectures (Jones et al., 2006; Cohen et al., 2010), but with significant challenges because of the difference listed below:

- ACT-R's restricts WM to being a fixed set of buffers, which leads to storing and retrievals of intermediate results from DM. In contrast, in Soar, intermediate results are maintained in WM, simplifying internal reasoning.
- In ACT-R, the primitive deliberative act involves the selection and firing of a single rule, whereas in Soar, it is the selection and application of an operator via a run-time composition of multiple rules. Soar's use of operators leads to impasse-driven subgoaling that supports metareasoning, planning, and other aspects of complex cognition that are more difficult to achieve in ACT-R.

- In Soar, DM is split into semantic memory and episodic memory. Episodic memory lets an agent easily access its past experiences and then reason about them which is much more challenging in ACT-R. This is the least fundamental difference in terms of overall design.

The organization of WM is perhaps the most fundamental difference between the two because it requires more deliberate reasoning to manage intermediate results. Conversely, one could imagine combining Soar's representations of operators with a constrained WM, possibly with an additional buffer that can represent impasse and substate information, providing for at least one level of metareasoning. Whether that is sufficient to enable the types of complex cognition found in Soar, and whether it is true to human processing constraints, are open questions.

Although there are important differences, and many other similarities, the most striking result of this analysis to me is the small number of different functional types of data (agent data, agent metadata, and module status data) and the consistency in how they interact in ACT-R and Soar to support both reasoning and learning. In both architectures, the flow of information among these types is the same. Agent data is represented in WM, PM, and DM, with different types of metadata for the agent data in each of those modules. However, agent metadata cannot be modified by agent data directly, and to date, cannot be tested by agent data (but see below). Module status data is read-only – it can be tested by agent data but cannot be modified. Importantly, it provides an agent with its only source of information about the current state of module processing. All other data available to the agent is about its environment, retrieved from memory, or produced during reasoning. Module status data provides access to the agent's own processing, providing an entry to metareasoning. Finally, the multiplicity of learning mechanisms use different combinations of agent data and metadata to either update metadata or create new agent data structures, but the learning mechanisms in the two architecture are very similar in data they use and the structures they create or modify. An interesting question is whether other architectures have these specific types of data and the corresponding interactions among them.

## 6.1 Data, Metadata, and Meta-process Data

In both ACT-R and Soar, metadata are critical for decision making, knowledge retrieval, and learning. They focus attention on the most relevant elements in PM and DM. Meta-process data, uniquely provided by module status data, provide internal feedback on processing and a potential entry for metacognition.

As observed earlier, both architectures maintain strict boundaries between agent data and agent metadata so that metadata are computed as a side effect of agent-data processing, and agent data cannot test or modify metadata. The meaning of metadata is thus fixed by the architecture, whereas the meanings of agent data are flexible and open to change through experience and learning new agent data. Moreover, behavior cannot be directly conditioned on metadata but only indirectly, such as through monitoring which chunk is retrieved given a specific context.

But is this strict separation necessary? Can metadata be available in WM and testable by PM without disrupting other commitments in these cognitive architectures? If metadata is available, it would require that the agent be imbued with, or learn, knowledge for interpreting the meaning of its metadata. As a possible example, the activation of a retrieved WM chunk could be available in the module buffer, providing some indication of confidence in the retrieval. Conversely, should agent data be able to modify metadata, such as through the actions of a PM element? This change is more problematic as agent data would interfere with the processes that compute and maintain metadata. Currently, agent data can modify metadata only *indirectly* through its other processing, such as rehearsal to boost the activation, where it relies on architectural mechanisms for changing activation. Thus, it may make sense to allow agent data to test metadata but not modify it, much as how meta-process data is handled.

In ACT-R and Soar, there are a few sources of data and metadata. Data in WM are derived from perception, agent data stored in memories, and module status data, whereas metadata are computed based on the processing of agent data. Are there other potential sources of data and metadata currently missing from ACT-R and Soar? Below are possible additional categories inspired by data types found in humans and by prior research on extending ACT-R and Soar.

1. <u>Memory and processing appraisals</u>. Humans can judge many properties of their memories and processing, such as familiarity in a belief, feelings about ease of processing; feelings of knowing such as tip-of-the-tongue, judgments of difficulty in learning, and judgments of remembering versus knowing. Some of these might be implemented as module status data for declarative memory. For example, feeling of knowing could be the status for a failed retrieval where the retrieval process has reason to believe a retrieval is possible. Remembering versus knowing could be based on the declarative memory buffer from which a result was retrieved (assuming separate semantic and episodic memories and buffers). Judgments of difficulty in learning are trickier and could require retrieval of similar past experiences where the agent explicitly evaluated the difficulty generalizes it to new situations.

2. <u>Judgments of passage of time.</u> In ACT-R, a module can be used that measures and estimates short time durations, but neither ACT-R nor Soar have mechanisms for judging longer-term time scales. Although Soar's episodic memory provides access to distant memories, explicit timing information and its use are still open research problems.

3. <u>Innate appraisals</u> (Marsella & Gratch, 2009). It is hypothesized that there are innate appraisals that are precursors to emotion. These appear to be properties of agent's overall processing state, often related to an agent's needs and goals, as well as to other overall properties of the situation, such as surprise. Important issues include how these are computed, how they influence other agent processes, such as memory retrieval, decision making, and learning (Marinier, Laird & Lewis, 2009), either indirectly as metadata or directly through elements in WM in some summary of the agent's overall embodiment.

4. <u>System-wide moderators.</u> Physiological moderators can have systemic effects on the cognitive system, including influencing attention, decision making, memory retrieval, and learning. Examples include fatigue, arousal, and stress. Chemicals such as dopamine, serotonin, and caffeine also have systemic effects. Researchers in ACT-R have examined many moderators, including fatigue (Gross et al., 2006) and other physiology factors (Darcy et al., 2015).

## 6.2 Future Work

A next step is to apply this analysis to similar cognitive architectures such as EPIC (Kieras & Meyer, 1997) and LIDA (Franklin et al., 2016). A more challenging analysis will be to examine architectures that use neural/graphical representations for memory elements, such as Sigma (Rosenbloom et al., 2016) and Spaun (Eliasmith et al., 2012). In these architectures, the data and metadata are intermixed, with symbolic structures implicit in the representations (Rosenbloom, et al., 2019). On promising hypothesis is that the structure of the *processing* of data and metadata will be similar, as was observed for Sigma in the Common Model paper (Laird et al. 2017), but more analysis is needed.

### Acknowledgments

Thanks to John Anderson, Dan Bothell, Nate Derbinsky, Steven Jones, Pat Langley, Christian Lebiere, Peter Lindes, Paul Rosenbloom, Bryan Sterns, and Andrea Stocco for comments on previous drafts, as well as the ACS reviewers.

This work was supported by AFOSR under Grant Number FA9550-18-1-0168 and the Office of Naval Research under Grant Number F048875-093099. The views and conclusions contained in this document are those of the authors and should not be interpreted as representing the official policies, either expressed or implied, of the Department of Defense, AFOSR, or Office of Naval Research. The U.S. Government is authorized to reproduce and distribute reprints for Government purposes notwithstanding any copyright notation hereon.